# Switchable Lightweight Anti-Symmetric Processing (SLAP) with CNN Outspeeds Data Augmentation by Smaller Sample – Application in Gomoku Reinforcement Learning

Chi-Hang Suen
City, University of London
London, UK
chi.suen@city.ac.uk

Eduardo Alonso
City, University of London
London, UK
e.alonso@city.ac.uk

*Abstract*—To replace data augmentation, this paper proposed a method called SLAP to intensify experience to speed up machine learning and reduce the sample size. SLAP is a model-independent protocol/function to produce the same output given different transformation variants. SLAP improved the convergence speed of convolutional neural network learning by 83% in the experiments with Gomoku game states, with only one eighth of the sample size compared with data augmentation. In reinforcement learning for Gomoku, using AlphaGo Zero/AlphaZero algorithm with data augmentation as baseline, SLAP reduced the number of training samples by a factor of 8 and achieved similar winning rate against the same evaluator, but it was not yet evident that it could speed up reinforcement learning. The benefits should at least apply to domains that are invariant to symmetry or certain transformations. As future work, SLAP may aid more explainable learning and transfer learning for domains that are not invariant to symmetry, as a small step towards artificial general intelligence.

*Keywords—data augmentation, convolutional neural network, symmetry invariant, group transformation, data preprocessing, SLAP, reinforcement learning*

I. INTRODUCTION

*A. Problem*

Convolutional neural network (CNN) is now the mainstream family of models for computer vision, thanks to its weight sharing mechanism to efficiently share learning across the same plane by so-called kernels, achieving local translational invariance. But CNN is not reflection and rotation invariant. Typically, it can be addressed by data augmentation to inputs by reflection and rotation if necessary, but the sample size would increase substantially. [1] criticised CNN that it could not learn spatial relationships such as orientation, position and hierarchy and advocated their novel capsule to replace CNN. [2] improved capsule using routing by agreement mechanism and outperformed CNN at recognising overlapping images, but they also admitted that it tended to account for everything in the structure. This implies capsule is too heavy in computation. Inspired by the idea of capturing orientation information in capsule network [2], this paper proposed a novel method called Switchable Lightweight Anti-symmetric Process (SLAP), a protocol to produce the same output given different transformation variants, with the main research question: can symmetry variants be exploited directly by SLAP to improve and combine with CNN for machine learning?

Very often, we know in advance if a certain machine learning task is invariant to certain types of transformation, such as rotation and reflection. E.g. in Gomoku, the state is rotation (perpendicularly) and reflection (horizontally and vertically) invariant in terms of winning probability, and "partially" translation invariant. Symmetry is often exploited by data augmentation for deep learning. But this greatly increases the dataset size if all symmetry variants are included – e.g. there are 8 such variants for each Gomoku state. SLAP was invented in this paper to avoid such expansion (see I.B).

On the other hand, reinforcement learning is notorious for lengthy training time and large sample size required. Data augmentation may help improve performance in reinforcement learning, but it would increase the sample size. This research tried to kill two birds by one stone, SLAP, by applying with CNN in reinforcement learning (of Gomoku), challenging the widely used practice of data augmentation, aiming at reducing the sample size and improving the learning speed.

*B. Switchable Lightweight Anti-symmetric Process (SLAP)*

SLAP is a model-independent protocol and function to always produce or choose the same variant regardless of which transformation variant (by specified symmetry) is given, and if required also output the corresponding transformation. It can be used upon any function or model to produce outputs that are invariant regarding specified symmetric properties of the inputs. If some types of the outputs are not invariant but follow the same transformation, the corresponding transformation information from SLAP may be used to transform these outputs back. It can be viewed as standardization of symmetry, as opposed to standardization of scale. After processing, symmetric variants are filtered out – that's why it is named 'anti-symmetric process'. Ironically, with this anti-symmetric process, the function or model (e.g. CNN) to be fed would look as if it is symmetric with regard to whichever the symmetry variant is the input, and the same output is produced. It is a novel method to exploit symmetry variants in machine learning without increasing the number of training samples by data augmentation. The motivation is to concentrate experience to speed up learning, without enlarging the sample size by data augmentation. See details in III.A.

## C. Gomoku

Gomoku, or Five in a Row, is a 2-player board game, traditionally played with Go pieces (black and white stones) on a Go board (19x19), nowadays on 15x15 board. For experiments in this research, mini board 8x8 was used instead to save computation, and the rule of freestyle version was adopted:

- Black (first) and white place stones of his colour alternately at an unoccupied intersection point.
- Winner: first one to connect 5 stones of his colour in a straight line (horizontal, vertical or diagonal).
- Draw happens if the board is full without a winner.

Gomoku was chosen to demonstrate the benefit of SLAP because:

- Gomoku has huge number of state representations ($3^{225} \approx 2 \times 10^{107}$), justifying the use of neural network.
- Gomoku is rotation and reflection invariant, but only "partially" translation invariant, so ideal to test different transformations.
- Gomoku is Markov Decision Process, meeting a basic assumption of reinforcement learning.
- [3] and [4] showed a general effective reinforcement learning algorithm for board games and Gomoku is simple to implement.

## II. BACKGROUND

### A. CNN

CNN (convolutional neural network) has been widely used for computer vision, but it is known that CNN is weak to deal with changes by rotation or orientation unless with much larger sample size by data augmentation. To address this problem, [1] proposed that neural network should make use of their then novel capsule, learning to recognize an implicitly defined visual entity and output probability of its existence and instantiation parameters such as pose; they showed that a transforming auto-encoder could be learnt to force the output (which is a vector instead of scalar) of a capsule to represent an image property that one might want to manipulate. [2] showed that a discriminatively trained, multi-layer capsule system achieves state-of-the-art performance on MNIST and was considerably better than CNN at recognizing highly overlapping digits, using the so-called routing by agreement mechanism, and yet [2] admitted that one drawback was the tendency of capsule to account for everything in an image. It implies that the capsule might be too "heavy" for computation and so a lightweight method is required. On lightweight capsule, DSC-CapsNet was proposed as lightweight capsule network, which focused on computing efficiency and reducing number of parameters [5]; [6] proposed dense capsule network with fewer parameters – neither had structure similar to SLAP. The capsule network with routing by agreement algorithm was proved by [7] not to be a universal approximator, i.e. not fit to all kinds of problems. As such, this research did not attempt to replace CNN by capsule, but simply created SLAP to combine with CNN. Instead of forcing the output to represent certain transformation information (e.g. orientation angle), SLAP forces the input of different variants (e.g. different rotation angle) to give the same output variant (and output the transformation information e.g. angle, if needed). Nevertheless, the invention of SLAP was inspired by [1] & [2] trying to address the weakness of CNN. On symmetric CNN, [8] proposed to impose symmetry in neural network parameters by repeating some parameters and achieved 25% reduction in number of parameters with only 0.2% loss in accuracy using ResNet-101, a type of CNN; but unlike SLAP, symmetry was not imposed in the inputs.

### B. Groupoid in Gomoku

There are different Gomoku states of the same groupoid (see Fig. 1), which means having local symmetry but not necessarily global symmetry of the whole structure [9]. Groupoid is more challenging than symmetry or group, as some groupoids may not have the same status, e.g. see Fig. 1. But the potential for learning is huge as there are many more variants, e.g. 156 variants just by translation in Fig. 1.

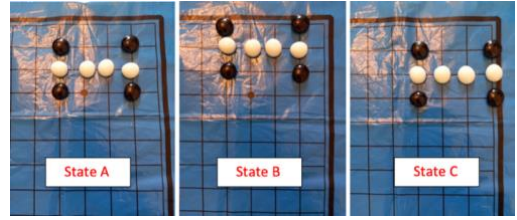

Fig. 1. Gomoku groupoid. Black can stop white win in C, but not in A or B.

### C. AlphaGo Zero / AlphaZero

For reinforcement learning of Gomoku in this research, the baseline algorithm was chosen to follow that of AlphaGo Zero [3] and Alpha Zero [4] because domain knowledge was not required. The algorithm was concisely summarized by [10] as follows:

*1) Neural Network*

The neural network feature extractor is a type of CNN. It takes state $s_t$ as input and yields value of state $v_\theta(s_t) \in [-1, 1]$ and policy $\vec{p_\theta}(s_t)$ as probability vector over all possible actions. It has the following loss function (excl. regularization terms):

$$\text{loss} = \sum_t (v_\theta(s_t) - z_t)^2 - \vec{\pi}_t . \log(\vec{p_\theta}(s_t)) \quad (1)$$

where $z_t$, $\vec{\pi}_t$ are final outcome {-1,0,1} and estimate (to be discussed below) of policy from state $s_t$ respectively, with 1, 0, -1 representing win, draw, lose respectively for current player.

*2) Monte Carlo Tree Search (MCTS) as policy improvement operator*

At each node, action is chosen by maximizing U(s, a), the upper confidence bound of Q-value Q(s, a):

$$U(s, a) = Q(s, a) + C * P(s, a) * \frac{\sqrt{\sum_b N(s,b)}}{1+N(s,a)} \quad (2)$$

where N(s, a) is no. of times taking action a from state s in MCTS simulation, P(s, .) is $\vec{p_\theta}(s)$, and the policy estimate of probability is improved by using (3):

$$\vec{\pi}_t = N(s, .) / \sum N(s, b) \quad (3)$$

When a new node (not visited before from parent node) is reached, instead of rollout, the value of new node is obtained from neural network and propagated up the search path. Unless the new node is terminal, the new node is expanded to have child nodes.

*3) Self-play training as policy evaluation operator*

In each turn, a fixed number of MCTS simulations are conducted from the state $s_t$, and action is selected by sampling from the policy estimate of probabilities improved by MCTS, thus generating training sample data. At the end of an iteration, the neural network is updated by learning from the training sample data.

The evaluation metric would be based on winning and drawing percentages of the AI against an independent evaluation agent. There are differences among AlphaGo Zero [3] and Alpha Zero [4], see Table I:

TABLE I.  DIFFERENCES BETWEEN ALPHAGO ZERO AND ALPHAZERO

|  | **AlphaGo Zero** | **AlphaZero** |
| --- | --- | --- |
| Pitting models | Yes, model with new weights plays against previous one; new weights are adopted only if it wins 55% or above | No, always use new weights after each iteration of neural network learning |
| Symmetry | Data augmentation by rotation and reflection to increase sample size by 8 times for training; transform to one of 8 variants randomly in self-play for inference | Not exploited, as it is intended for generalization |
| Action in self-play | Sampled proportional to visit count in MCTS in first 30 moves, then selected greedily by max visit count (asymptotically with highest winning chance) in MCTS | Sampled proportional to visit count in MCTS |
| Outcome prediction | Assume binary win/loss, estimate & optimise winning probability | Also consider draw or other outcomes, estimate & optimise expected outcome |

*D. Other Related Works, Symmetry and AGI*

[11] incorporated symmetry into neural network by creating symmetry (of specific type) invariant features, but no implementation or idea similar to SLAP was used. Studies have shown rotation-based augmentation performed better than many other augmentation techniques [12]. The type of data augmentation used as baseline in this research was rotation and reflection based (also the type used by AlphaGo Zero[3]). The novelty lies in the fact that SLAP is opposite to the practice of data augmentation – decreasing the variety of variants in the data instead for machine learning, though also exploiting symmetry.

Symmetry is one of the natures of the real world. Animals can detect the same object or the same prey being moved (translated), or even rotated after being slapped (the novel method was deliberately abbreviated as SLAP). Recognising symmetry can also speed up learning patterns, a typical trick used for playing some board games. To facilitate research exploiting symmetry in machine learning, [13] connected symmetry transformations to vector representations by the formalism of group and representation theory to arrive at the first formal definition of disentangled representations, expected to benefit learning from separating out (disentangling) the underlying structure of the world into disjoint parts of its representation. Upon this work, [14] showed by theory and experiments that Symmetry-Based Disentangled Representation Learning (SBDRL) could not only be based on static observations: agents should interact with the environment to discover its symmetries. They emphasized that the representation should use transitions rather than still observations for SBDRL. This was taken into account for designing the Gomoku representation for reinforcement learning in this research.

One may expect that an artificial general intelligence (AGI) system, if invented, should be able to learn unknown symmetry. Researchers have worked on this, for example [15] proposed learning unknown symmetries by different principles of family of methods. But it is equally important to learn by exploiting symmetry more effectively. For example, if an AGI system can interpret the rules of Gomoku and realize from the rules that Gomoku is reflection and rotation invariant, it should directly exploit such symmetry instead of assuming symmetry is unknown. Ideally, such exploitation should be switched on easily if one wishes, and hence the term 'switchable' in SLAP, which can be used upon any function or model. If transfer learning in CNN is analogous to reusing a chair by cutting the legs and installing new legs to fit another, such 'switchable learning' in SLAP is analogous to turning the switch of an adjustable chair to fit certain symmetries. Such kind of 'switch' in design can also help AI be more explainable and transparent, and more easily reused or transferred, while an AGI system should be able to link and switch to different sub-systems easily to solve a problem. SLAP can also reduce memory required. For example, AlphaGo Zero used a transposition table [3], a cache of previously seen positions and associated evaluations. Had SLAP been used instead of data augmentation, such memory size could be reduced by a factor of 8, or alternatively 8 times more positions or states could be stored. Indeed memory plays an important role in reinforcement learning as well by episodic memory, an explicit record of past events to be taken as reference for making decisions, improving both sample efficiency and speed in reinforcement learning as experience can be used immediately for making decisions [16]. It is likely that an AGI system would, just like human, use memory to solve some problems rather than always resort to learning from scratch. And in the real world, a continuous space, there can be much more than 8 equivalent variants. Recently, [17] suggested that symmetry should be an important general framework that determines the structure of universe, constrains the nature of natural tasks and consequently shape both biological and artificial general intelligence; they argued that symmetry transformations should be a fundamental principle in search for a good representation in learning. Perhaps SLAP may contribute a tiny step towards AGI, by shaping input representations directly by symmetry transformation. Note that SLAP can be used upon any function or model, and even if some (types) of the outputs are not invariant but follow the same transformation, these may be broken down and use the transformation information output from SLAP to make appropriate transformation back later for these parts only. A little kid often mistakes *b* for *d* at the beginning of learning alphabets, and it appears that human learning types of objects by vision might naturally assume symmetry first and then learn non-symmetry later. If a machine learning problem is to be split into stages or parts by specified symmetry as a guide, SLAP might help by wrapping certain parts of a function or neural network model.

III. METHODS

*A. SLAP*

SLAP forces the input of different variants (e.g. different rotation angle) to give the same output variant (and output the transformation information e.g. angle, though not necessarily used). There can be multiple ways to achieve this. For rotation and reflection variants of Gomoku states, one way to implement this is simply flattening the pixels of 8 variants to 8 lists, compare the lists and always choose the largest. Below (Fig. 2) was the algorithm used for SLAP in dealing with rotation and reflection variants of Gomoku states, but the concept may be applied to other symmetries as well.

**Algorithm SLAP**

| |
|---|
| 1: Generate symmetry variants of input, store required transformation |
| 2: Convert each variant to a list |
| 3: Compare each list and find the 'largest' list |
| 4: **return** the 'largest' *variant* & required transformation of the variant |

Fig. 2. SLAP algorithm. Positive large data cluster towards top left.

If the image/state has multiple input channels or planes in one sample, the first channel/plane is compared first by list comparison.

SLAP was implemented by numpy instead of torch tensor for faster speed, because numpy uses view for rotation and reflection. The output variant replaced the input state when SLAP was applied in training. During inference time, output action probabilities from neural network would be transformed back using the transformation information (rotation & reflection) from SLAP.

*1) Invariance*

Denote s, t = slap($x_i$), where slap is SLAP function in pythonic style, s is the symmetry (of certain group G, with n symmetry variants for each state) variant and t is corresponding transformation information. Given property of slap, for all i∈$N_{<=n}$,

$$s, t = slap(x_1) = slap(x_2) = \ldots = slap(x_n) \quad (4)$$

Denote s = slap($x_i$)[0], t = slap($x_i$)[1], the pythonic expression to capture first and second return variables of a function respectively. Denote h(slap($x_i$)[0]) as $h^{slap}(x_i)$ for any function h. Given an arbitrary function y = f(x),

$$y = f^{slap}(x_i)$$

By definition, ⇒ y = f(slap($x_i$)[0])

Using (4), ⇒ y = f(s) for all i (5)

∴ y = $f^{slap}(x_i)$ is invariant with respect to i (i.e. symmetry of group G).

When f is the neural network, the composite function resulting from the neural network, $f^{slap}$, is invariant to symmetry (of group G).

*2) Differentiability*

SLAP was not applied to intermediate layers of neural networks for Gomoku, so its differentiability was not required in this research. Approximation would be required to make it differentiable.

*3) Groupoid and SLAP-CC*

As Gomoku is only 'partially' invariant to translation, it is also interesting to experiment with translation variants, which are considered to be groupoid instead of group as they are symmetric locally but not necessarily symmetric globally. There can be many more translation variants than rotation and reflection variants, see II.B. To save computation, another algorithm (crop and centre) was used to 'standardize' translation variants. It was denoted as SLAP-CC in the below to emphasize that it shared the same general idea as SLAP, but just different way for implementation. Denoted as *cc* in the code.

The algorithm of SLAP-CC, shown in Fig. 3, would concentrate experience around the centre, as input variant was centred to become output variant. If it could not be exactly centred, the algorithm would make it slightly lean to top left.

**Algorithm SLAP-CC**

| |
|---|
| 1: Find non-empty min & max indices by row & column in input image |
| 2: r_shift = (no. of rows –1 –min row index –max row index)//2 |
| 3: c_shift=(no. of columns –1 –min column index –max column index)//2 |
| 4: **return** numpy.roll(image, (r_shift, c_shift), axis=(-2, -1)) |

Fig. 3. SLAP-CC algorithm. Non-zero data cluster towards centre.

Note that since Gomoku is not completely invariant to translation, SLAP-CC was used to add information as additional planes instead, as opposed to replacing the input state when SLAP was applied. 2 planes representing stones of different colours (current and opponent players respectively) centred together by SLAP-CC, followed by 2 planes representing original indices of vertical and horizontal positions respectively (scaled linearly to [1, -1]) were added along with original 4 planes in Gomoku state representation (see III.B). The scaled position indices for whole plane were to give neural network a sense of original positioning.

*B. Representation of Gomoku*

In this research, the representation of Gomoku followed the style of AlphaGo Zero / AlphaZero, with simplification and taking [14] into account for representation design.

For each Gomoku state, there were 4 planes representing current player stones, opponent stones, last action and current colour respectively by one-hot-encoding. See Fig. 4 for a typical Gomoku state in this research, which used simplified board size 8x8 instead.

```
00000000   00000000   00000000   11111111
00000000   00000000   00000000   11111111
00000000   00000000   00000000   11111111
00010000   00000000   00000000   11111111
00010000   00001000   00000000   11111111
00000000   00010000   00010000   11111111
00000000   00000000   00000000   11111111
00000000   00000000   00000000   11111111
Current player  Opponent player  Last action  Current Colour
```

Fig. 4. Gomoku state representation example at time t = 4.

For labels, probabilities of a move over all positions were represented by 8x8 flattened vector. Final outcome (value) of current player was represented by 1, 0, -1 respectively for win, draw, lose.

*C. SLAP in Gomoku Reinforcement Learning*

SLAP was used to pre-process states for network training and inference. Transformation information from SLAP was only used in network inference to convert probabilities (not estimated outcome) back to corresponding game board positions for MCTS to improve probabilities of actions, which were used as sampling probabilities to make a move in self-play (but greedy in evaluation). See Fig. 5.

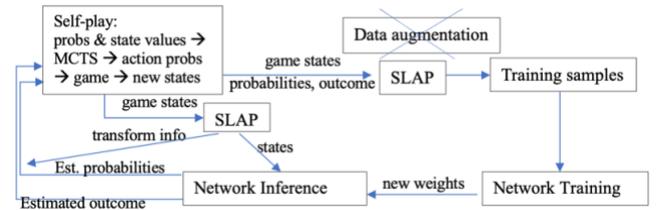

Fig. 5. SLAP used in Gomoku reinforcement learning.

For SLAP-CC, it was applied at the same place as SLAP in the above flow chart, but data augmentation was kept instead of being replaced and no transformation information was used to transform probabilities output of the network. See methods in III.A.3.

*D. Testing Benefits for Neural Network Learning*

To decouple from reinforcement learning dynamics, synthetic states of Gomoku were created for testing neural network learning with SLAP vs with typical data augmentation (by rotation and reflection), the latter of which had 8 times the number of training samples. Self-play was not involved in this testing.

Synthetic states were generated by first creating states each with only 5 stones connected in a straight line (i.e. win status) for all combinations for current black player, then removing one stone (to be repeated with another stone 5 times to create 5 different states) and randomly adding 4 opponent stones to become one about-to-win state. Together these were one set of 480 about-to-win states. Different sets could be created since white stones were merely random. Each set was mixed with 1000 purely random states, also with 4 stones for each player. 8 mixed sets were created, i.e. 11,840 states. 15%, i.e. 1,776 were reserved for validation test.

Labels were assigned as follows: if there were one or more choices to win immediately (include some purely random states, though the chance would be very remote), the value of state would be labelled as 1 and the wining position(s) would be labelled with probability of move = 1/no. of winning positions, while others were labelled 0; otherwise the value of state would be labelled as 0 and the probability of move for each available position would be random by uniform distribution, normalizing and summing to 1.

Neural networks (see Appendix) with SLAP vs with data augmentation would learn from training samples of states and labels to predict labels of validation data given the input states. Validation loss and its speed of convergence would be the key metrics.

First, at preliminary stage, for each set of hyperparameters the neural network ran 1000 iterations each with batch size 512 sampled from training samples of size 10,064 and 80,512 respectively for neural networks with SLAP and neural networks with data augmentation. Sampled with replacement, same as during reinforcement learning. There were 2400 combinations of hyperparameters by grid search, shown in Table II:

TABLE II. HYPERPARAMETERS TESTED AT PRELIMINARY STAGE OF CNN LEARNING

| Hyperparameter | Tested values | Remarks |
|---|---|---|
| use_slap | True, False | False: data augmentation instead of SLAP |
| extra_act_fc | True, False | True: add extra layer (size 64) to action policy |
| L2 | $10^{-3}, 10^{-4}, 10^{-5}$ | weight decay of optimizer |
| Num_ResBlock | 0, 5, 10, 20 | no. of residual blocks |
| SGD | True, False | False: Adam optimizer |
| lr | $10^{-1}, 10^{-2}, 10^{-3}, 10^{-4}, 10^{-5}$ (learning rate) | |
| dropout | 0, 0.1, 0.2, 0.3, 0.4 | |

If Num_ResBlock > 0, the residual blocks replaced the common CNN layers and added a convolutional layer of 256 filters (3x3 kernel, stride 1, padding 1, no bias, ReLU activation) as the first layer. No autoclip [18] in optimizer, unlike reinforcement learning.

At stage 2, selected models from previous stage would run for 10,000 iterations instead of 1,000 iterations, with losses recorded every 10 iterations.

*E. Testing Benefits for Reinforcement Learning*

The baseline algorithm of Gomoku reinforcement learning followed AlphaGo Zero/AlphaZero (see II.C). Among their differences, the baseline algorithm in this research followed the better version, and thus followed AlphaZero except on symmetry exploitation. Like AlphaGo Zero, the baseline exploited symmetry by data augmentation to increase no. of training samples by 8 times, but random transformation was not done in self-play. Autoclip to gradients [18] was added in the optimizer for stable learning.

Reinforcement learning required much more computation than neural network learning, so to save computation, the same neural network will be used and the testing of hyperparameters would be based on best models in neural network learning by synthetic Gomoku states, with some deviations to be tested by grid search.

Stage 1: each of 240 models were trained by self-play of 250 games. Data buffer size: 1,250 and 10,000 for SLAP and non-SLAP models respectively, both roughly equivalent to storing latest 60 games.

Stage 2: selected models were trained by self-play of 5000 games. With more games arranged for training, larger data buffer size could be used. So, data buffer size was increased to 5000 and 4000 respectively for SLAP and non-SLAP models, roughly equivalent to storing latest 250 games. To align with stage 1 testing initially, the initial data buffer size was kept same as stage 1 for first 1000 games. This also got rid of initial poor-quality game state data quickly. Learning rate multiplier was used to adaptively decrease learning rate by half if validation loss increased beyond 3-sigma limit, measured every 100 games.

Evaluation: Independent agent(s), also called evaluation agent or evaluator, was built by pure Monte Carlo Tree Search (MCTS) with random policy to play against the trained AI. The strength of a pure MCTS agent depends on no. of playouts (aka. simulations) in each move. To facilitate observation of growing strength, multi-tier evaluation was built by playing 10 games against each of 3 pure MCTS agents (30 games total), each with 1000, 3000, 5000 playouts respectively. Overall winning rate (tie counted as half win) against them would be the key metrics for reinforcement learning. It was often either a win or loss, and seldom a tie. Assuming that a tie could be neglected, especially after counting tie as half win, it simplified as Bernoulli distribution with standard deviation approximated by $\sqrt{p(1-p)/30}$ to calculate confidence interval, where 30 is the number of trials in each evaluation.

*F. Code Implementation*

The part regarding AlphaZero was upgraded from [19]. Details of implementation and code repository: https://github.com/chihangs

## IV. RESULTS

### A. Impact on Neural Network Learning

*1) SLAP vs baseline (data augmentation)*

The best few SLAP and baseline models converged to loss around 2.81 (difference < 0.01), all without residual blocks. 3 SLAP models (denoted as s0_...) and 3 baseline models (denoted as n0_...) were selected and their losses were plotted in Fig. 6, where each model had Adam optimizer, same learning rate 0.001, no dropout, no residual blocks, but different values of L2 ($10^{-3}$, $10^{-4}$, $10^{-5}$).

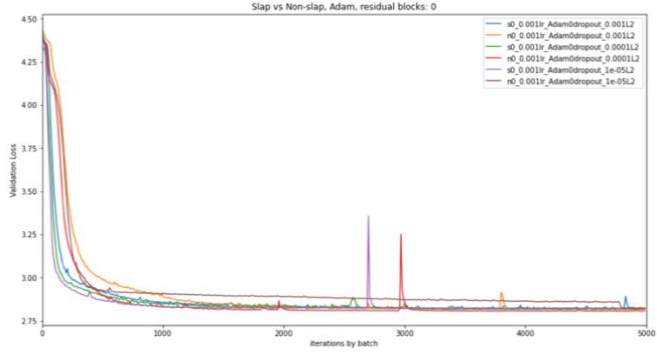

Fig. 6. Validation losses of SLAP and baseline models.

Above 6 models were repeated 3 more times to calculate average time (by no. of iterations) for convergence. SLAP speeded up the convergence by 95.1% and 71.2% measured by validation loss reaching 3.0 and 2.9 respectively, 83.2% in average.

*2) Testing sample size*

Holding validation dataset unchanged, the training data sample size was reduced by holding out some samples to match required size, using models with L2=$10^{-4}$ from Fig. 6. SLAP models converged when sample size was 5032 or above, but they were more vulnerable to decreasing no. of training samples and failed to converge when the sample size decreased to 2516 or below, while their baseline counterpart models (8 times the sample size) still converged.

*3) SLAP-CC vs baseline (data augmentation)*

SLAP-CC (see III.A.3) was added to the 3 best baseline models from Fig. 6. Validation losses of SLAP-CC converged to around 2.8 for all 3 values of L2, like its baseline counterparts. Experiments were repeated 3 more times to calculate average time (by no. of iterations) to converge. The time for validation loss to reach 3.0 and 2.9 both worsened by 30.7% in average for SLAP-CC.

### B. Impact on Neural Network Learning

*1) SLAP vs baseline (data augmentation)*

The best SLAP model had highest winning rate 86.7%, equivalent to winning 26 games out of 30. 95% confidence interval was 86.7% +/- 12.2%, i.e. (74.5%, 98.9%). The best baseline model had highest winning rate 93.3%, equivalent to winning 28 games out of 30, with 95% confidence interval 93.3% +/- 8.9%, i.e. (84.4%, 100%).

The best SLAP and baseline models had similar winning rates, by confidence intervals. If winning rate of two thirds (66.6%) is used as benchmark for this three-tier evaluation, both took 1000 games to achieve or surpass this. However, non-SLAP took 1250 games only to first achieve winning rate of 86.6%, while SLAP took 3000 games.

SLAP spent 0.761 second per move in self-play, 10.8% more time than baseline (only 5% more in a separate speed-optimizing version). SLAP tended to decrease learning rate multiplier more frequently, implying more frequent significant increase of validation loss.

*2) Testing buffer size*

Best models of SLAP and non-SLAP were repeated but with smaller data buffer size of only 1,250 and 10,000 respectively throughout whole reinforcement learning. Similar to stage 2, above models were trained by 5000 games. With fewer data in buffer, the highest winning rate achieved for SLAP model was only 73.3%, below the corresponding confidence interval. The highest winning rate achieved for non-SLAP model was only 83.3%, below the corresponding confidence interval. So, it harmed reinforcement learning when data buffer was too small, and it was good decision to use larger data buffer at stage 2.

*3) SLAP-CC vs baseline (data augmentation)*

SLAP-CC was tested by same configurations as best baseline model from IV.B.1 but adding information from SLAP-CC and scaled position indices as extra input feature planes. The new model also ran for 5000 games. See methods in III.A.3 and III.C. Learning rate multiplier did not change throughout training. The best winning rate achieved for SLAP-CC model was 96.7%, slightly higher than the baseline, but within the confidence interval.

## V. DISCUSSION

Despite the widely use of data augmentation to increase the variety of transformation variants in samples to improve machine learning, we proved that using SLAP to decrease the variety could achieve the same performance of typical data augmentation with sample size reduced by 87.5% and faster by 83.2% in CNN network learning, and statistically the same performance for reinforcement learning with sample size reduced by 87.5%. The success could be explained by concentrating learning experience to certain regions when different variants were transformed by SLAP, implicitly sharing weights among variants. The proof of invariance (see III.A.1) after applying SLAP did not require the network to be CNN and it could be an arbitrary function, so the applicability of SLAP should not be restricted to CNN. While SLAP exploited only reflection and rotation symmetries in learning Gomoku, the general concept of SLAP and the proof of invariance could apply to other symmetries. As no domain specific features or knowledges (except symmetry) were used in SLAP, the benefits shown in the experiments should apply generally for domains that are symmetry invariant.

Shortcomings: in Gomoku reinforcement learning, SLAP tended to decrease learning rate multiplier more frequently, implying more frequent significant increase of validation loss. This instability could be caused by faster neural network learning. Note that AlphaGo Zero only dropped learning rate twice over 1,000,000 training steps in their planned schedule [3]. It might imply that SLAP would need quite different hyperparameters in reinforcement learning (as opposed to sharing the same hyperparameters of baseline models in the neural network learning experiment), and more or better searches of hyperparameters for reinforcement learning would be required, though it was constrained by computation resources. Another plausible explanation for not speeding up reinforcement learning was the insignificant portion (~1%-2%) of neural network learning in the whole reinforcement training, implying that the time saved in neural network learning would be insignificant for the whole reinforcement learning in our chosen setting (which used a relatively simple CNN), and enough neural network learning iterations would have been allowed if hyperparameters were optimal.

Limitations: the results only applied to symmetry-invariant domain, and SLAP could be more vulnerable if the sample was too small (see IV.A.2). SLAP required 10.8% more time for self-play in IV.B.1, but the overhead would be insignificant if the simple CNN were replaced by a deep one. It was not yet proved to speed up reinforcement learning. Neither was it proved to be able to exploit groupoid patterns.

## VI. Conclusion and future work

SLAP could improve the convergence speed of neural network (CNN in the experiment) learning synthetic states in Gomoku by 83.2%, with only one eighth of training sample size of baseline model (data augmentation). Since no domain specific features or knowledges were used in SLAP, it should also benefit neural network learning generally for domains that are symmetry invariant, especially for reflection and rotation symmetries. As SLAP is model-independent, the benefits should apply to models beyond CNN. But it was not yet proved to speed up reinforcement learning, though it could achieve similar performance with smaller training sample size. Neither was it proved to exploit groupoid variants effectively.

As future work, SLAP may be applied in domains that are not fully symmetry invariant, by breaking down the neural network layers into two parts – first learning as if it were fully symmetry invariant. Or even split into stages by type of symmetries. Although SLAP is not directly differentiable, one workaround would be similar to that in transforming Gomoku action probabilities: given the transformation information as another input, transform the learned output back to corresponding original position, and then carry out necessary subsequent computations forward. This helps create more explainable stages and transfer learning. Another future work might be differentiable approximation of SLAP.

## Appendix

*Neural Network Architecture and Configurations*

The architecture and configurations used (unless otherwise stated):

Architecture: consisted of 3 common convolutional layers (32, 64, 128 filters respectively) each with 3x3 kernel of stride 1 and padding 1 with ReLU activation, followed by 2 action policy players and in parallel 3 state value layers. The input was 8 x 8 x 4 image stack comprising of 4 binary feature planes. The action policy layers had one convolutional layer with 4 filters each with 1x1 kernel of stride 1 with ReLU activation, followed by a fully connected linear layer to output a vector of size 64 corresponding to logit probabilities for all intersection points of the board. The state value layers had one convolutional layer with 2 filters each with 1x1 kernel of stride 1 with ReLU activation, followed by fully connected linear layer to a hidden layer of size 64 with ReLU activation, finally fully connected to a scalar with tanh activation. Dropout would be applied to all action policy layers and state value layers except output layers; not applied to common layers.

Optimizer: Adam with autoclip [18]
Batch size per optimisation step: 512 (2048 in [3])
Data buffer size: 10,000 for baseline, 1,250 for SLAP
No. of network optimisation steps per policy iteration: 10
No. of self-play games per policy iteration: 1
No. of playouts: 400 (1600 in [3], 800 in [4])
$C_{puct}$ (constant of upper confidence bound in MCTS) : 5
Temperature parameter: 1 (same as in [4])
Dirichlet alpha of noise: 0.3 (same as chess in [4])

Smaller batch size and no. of playouts per move in MCTS were used because Gomoku is less complex than Go. Dirichlet alpha was initially set at 0.3 because mini Gomoku (8x8 board) has same board size as chess and similar no. of available action choices per move.

## References

[1] G. E. Hinton, A. Krizhevsky and S. D. Wang, 'Transforming auto-encoders' in *International Conference on Artificial Neural Networks (ICANN)*, 2011.
[2] S. Sabour, N. Frosst and G. E. Hinton, 'Dynamic routing between capsules' in *31st Conference on Neural Information Processing Systems (NIPS 2017)*, Long Beach, CA, USA, Oct 2017.
[3] D. Silver et al., 'Mastering the game of go without human knowledge', *Nature,* 550, pp. 354– 359, Oct 2017.
[4] D. Silver et al., 'Mastering chess and Shogi by self-play with a general reinforcement learning algorithm', *Science*, Vol 362, Issue 6419, pp. 1140-1144, Dec 2018.
[5] S. Dan, X. Hu, Y. Zhou and S. Duan, 'Lightweight multi-dimensional memristive CapsNet' in *International Joint Conference on Neural Networks (IJCNN)*, Shenzhen, China, Jul 18-22 2021.
[6] K. Sun, X. Wen, L. Yuan and H. Xu, 'Dense capsule networks with fewer parameters', *Soft computing (Berlin, Germany)* (1432-7643), Vol 25 (Issue 10), pp. 6927, Apr 2021.
[7] D. Peer, S. Stabinger and A. Rodriguez-Sanchez, 'Limitations of capsule networks', *ScienceDirect (Pattern Recognition Letter)*, Vol 144, pp. 68-74, Apr 2021.
[8] X. S. Hu, S. Zagoruyko and N. Komodakis, 'Exploring weight symmetry in deep neural networks' 2018, *arXiv:1812.11027*.
[9] A. Vistoli, 'Groupoids: a local theory of symmetry', *Isonomia (Epistemologica)* 2011, 26, pp. 1–12, Sep 2011.
[10] S. Nair. 'A simple Alpha(Go) Zero tutorial' Stanford University. Accessed: 18 May 2022. [Online.] Available: http://web.stanford.edu/~surag/posts/alphazero.html
[11] D. Bergman, 'Symmetry constrained machine learning' *2019, arXiv:1811.07051v2*.
[12] K. Maharana, S. Mondal and B. Nemade, 'A review: data pre-processing and data augmentation techniques', *Global Transitions Proceedings*, vol. 3, Issue 1, pp. 91-99, Jun 2022.
[13] I. Higgins et al., 'Towards a definition of disentangled representations' *2018, arXiv:1812.02230*.
[14] H. Caselles-Dupré, M. Garcia-Ortiz and D. Filliat, 'Symmetry-based disentangled representation learning requires interaction with environments' in *33rd Conference on Neural Information Processing Systems (NeurIPS 2019), Vancouver, Canada, 2019*.
[15] F. Anselmi, G. Evangelopoulos, L. Rosasco and T. Poggio, 'Symmetry regularization', *The Centre for Brain, Mind and Machines (CBMM) Memo No. 63*, May 2017.
[16] M. Botvinick et al., 'Reinforcement learning, fast and slow' *Trends in Cognitive Sciences*, Vol 23, Issue 5, pp. 408-422, Apr 2019.
[17] I. Higgins, S. Racanière and D. Rezende, 'Symmetry-based representations for artificial and biological intelligence', *Frontiers in Computational Neuroscience*, vol. 16, 2022, doi: 10.3389/fncom.2022.836498.
[18] P. Seetharaman, G. Wichern, B. Pardo and J. L. Roux, 'AutoClip: adaptive gradient clipping for source separation networks' in *2020 IEEE 30th International Workshop on Machine Learning for Signal Processing (MLSP)*, Espoo, Finland, Sep 21-24 2020, doi: 10.1109/MLSP49062.2020.9231926.
[19] J. Song. 'An implementation of the AlphaZero algorithm for Gomoku (also called Gobang or Five in a Row)' github.com. Accessed: Jun-Sep 2022. [Online.] Available: https://github.com/junxiaosong/AlphaZero_Gomoku